\documentclass[sigconf]{acmart}
\AtBeginDocument{%
  \providecommand\BibTeX{{%
    \normalfont B\kern-0.5em{\scshape i\kern-0.25em b}\kern-0.8em\TeX}}}

\copyrightyear{2023} 
\acmYear{2023} 
\setcopyright{acmlicensed}\acmConference[WWW '23]{Proceedings of the ACM Web Conference 2023}{May 1--5, 2023}{Austin, TX, USA}
\acmBooktitle{Proceedings of the ACM Web Conference 2023 (WWW '23), May 1--5, 2023, Austin, TX, USA}
\acmPrice{15.00}
\acmDOI{10.1145/3543507.3583199}
\acmISBN{978-1-4503-9416-1/23/04}

\usepackage{times}
\usepackage{latexsym}

\usepackage{blindtext}

\usepackage{tikz}

\usepackage{tabularx,ragged2e}
\usepackage{ltablex}
\usepackage{makecell}

\usepackage{supertabular}

\usepackage{booktabs}
\usepackage{multirow}
\usepackage[utf8]{inputenc}

\usepackage{microtype}

\usepackage{xcolor}
\usepackage{soul}
\usepackage{threeparttable} 

\DeclareUrlCommand\url{\color{blue}}

\usepackage{microtype}

\usepackage{subcaption}
\usepackage{graphicx}
\usepackage{enumitem}
\usepackage{colortbl}

\begin{document}

\DeclareRobustCommand{\hlcyan}[1]{{\sethlcolor{yellow}\hl{#1}}}
\DeclareRobustCommand{\hlgreen}[1]{{\sethlcolor{brown}\hl{#1}}}
\newcommand{\blue}[1]{{\color{blue}#1}}
\newcommand{\cnote}[1]{{\color{blue} [\bf   #1]}}
\newcommand{\lee}[1]{{\color{red} [\bf {Dongwon:}  #1]}}
\newcommand{\thai}[1]{{\color{purple} [\bf {Thai:}  #1]}}
\newcommand{\todo}[1]{\sethlcolor{yellow}\hl{#1}}

\definecolor{HLcolor}{RGB}{124,18,18}
\sethlcolor{HLcolor!20}

\makeatletter
\newdimen\SOUL@dimen %new
\def\SOUL@ulunderline#1{{%
    \setbox\z@\hbox{#1}%
    \SOUL@dimen=\wd\z@ %new
    \dimen@i=\SOUL@uloverlap
    \advance\SOUL@dimen2\dimen@i %\dimen@ exchanged too
    \rlap{%
        \null
        \kern-\dimen@i
        \SOUL@ulcolor{\SOUL@ulleaders\hskip\SOUL@dimen}% new
    }%
    \unhcopy\z@
}}
\makeatother

\newcommand\MyDBox[2][HLcolor!20]{\sethlcolor{#1}\hl{#2}}

\newcommand{\centered}[1]{\begin{tabular}{l} #1 \end{tabular}}

\title{Do Language Models Plagiarize?}


%\author{
%    ~Jooyoung Lee\textsuperscript{1}, 
%    ~Thai Le\textsuperscript{2},
%    ~Jinghui Chen\textsuperscript{1}, 
%    ~Dongwon Lee\textsuperscript{1}
%    \\
%    \textsuperscript{1}The Pennsylvania State University\\
%    \textsuperscript{2}University of Mississippi   
%}

\author{Jooyoung Lee}
\email{jfl5838@psu.edu}
\affiliation{%
  \institution{Penn State University}
  \city{University Park}
  \state{PA}
  \country{USA}
}

\author{Thai Le}
\email{thaile@olemiss.edu}
\affiliation{%
  \institution{University of Mississippi}
  \city{Oxford}
  \state{MS}
  \country{USA}
}

\author{Jinghui Chen}
\email{jzc5917@psu.edu}
\affiliation{%
  \institution{Penn State University}
  \city{University Park}
  \state{PA}
  \country{USA}
}

\author{Dongwon Lee}
\email{dongwon@psu.edu}
\affiliation{%
  \institution{Penn State University}
  \city{University Park}
  \state{PA}
  \country{USA}
}

\renewcommand{\shortauthors}{Lee et al.}

\begin{abstract}
Past literature has illustrated that language models (LMs) often \emph{memorize} parts of training instances and reproduce them in natural language generation (NLG) processes. However, it is unclear to what extent LMs ``reuse'' a training corpus. For instance, models can generate paraphrased sentences that are contextually similar to training samples. In this work, therefore, we study three types of \emph{plagiarism} (i.e., verbatim, paraphrase, and idea) among GPT-2 generated texts, in comparison to its training data, and further analyze the plagiarism patterns of fine-tuned LMs with domain-specific corpora which are extensively used in practice. Our results suggest that (1) three types of plagiarism widely exist in LMs beyond memorization, (2) both size and decoding methods of LMs are strongly associated with the degrees of plagiarism they exhibit, and (3) fine-tuned LMs' plagiarism patterns vary based on their corpus similarity and homogeneity. 
%
%\lee{need to say a bit how the paper fits WWW context}
%
Given that a majority of LMs' training data is scraped from the Web \emph{without informing content owners}, their reiteration of words, phrases, and even core ideas from training sets into generated texts has ethical implications. Their patterns are likely to exacerbate as both the size of LMs and their training data increase, raising concerns about indiscriminately pursuing larger models with larger training corpora. Plagiarized content can also contain individuals' personal and sensitive information. These findings overall cast doubt on the practicality of current LMs in mission-critical writing tasks and urge more discussions around the observed phenomena.
%\textit{Data and source code will be shared after review is complete.}\lee{add the link} 
\textit{Data and source code 
%will be shared 
are available at \url{https://github.com/Brit7777/LM-plagiarism}.}

\end{abstract}

\begin{CCSXML}
<ccs2012>
   <concept>
       <concept_id>10010147.10010178.10010179.10010182</concept_id>
       <concept_desc>Computing methodologies~Natural language generation</concept_desc>
       <concept_significance>500</concept_significance>
       </concept>
 </ccs2012>
\end{CCSXML}

\ccsdesc[500]{Computing methodologies~Natural language generation}

\keywords{Language Models, Natural Language Generation, Plagiarism}



\maketitle

\newcolumntype{P}[1]{>{\centering\arraybackslash}p{#1}}

%\newcolumntype{C}{>{\Centering\arraybackslash}X} % centered "X" column
\begin{table*}[ht]
 \centering
{\footnotesize
\hfill{}
\begin{tabular}{P{0.09\textwidth}|P{0.41\textwidth}P{0.41\textwidth}}
%\begin{xtabular}{|p{0.48\textwidth}|p{0.48\textwidth}|}
\toprule
Type & Machine-Written Text & Training Text \\
\cmidrule(lr){1-3}
\multirow{3}*{Verbatim} & \MyDBox[yellow!30]{*** is the second amendment columnist for Breitbart news and host of bullets with ***, a Breitbart news podcast. {\color{olive}[...]}}  (\textit{\textbf{Author: GPT-2}}) &  \MyDBox[yellow!30]{*** is the second amendment columnist for Breitbart news and host of bullets with ***, a Breitbart news podcast. {\color{olive}[...]}} \\ 
\cmidrule(lr){1-3}
\multirow{6}*{Paraphrase} & \MyDBox[orange!30]{Cardiovascular disease, diabetes and hypertension significantly increased the risk of severe COVID-19}, and \MyDBox[orange!30]{cardiovascular disease increased the risk of mortality.} (\textit{\textbf{Author: Cord19GPT}}) & For example, \MyDBox[orange!30]{the presence of cardiovascular disease is associated with an increased risk of death from COVID-19} [14] ; \MyDBox[orange!30]{diabetes mellitus, hypertension, and obesity are associated with a greater risk of severe disease} [15] [16] [17] [18]. \\

\cmidrule(lr){1-3}
\multirow{9}*{Idea} & A system for automatically creating a plurality of electronic documents based on user behavior comprising: {\color{olive}[...]} and wherein \MyDBox[orange!30]{the system allows a user to choose an advertisement selected by the user for inclusion in at least one of the plurality of electronic documents}, \MyDBox[orange!30]{the user further being enabled to associate advertisement items with advertisements for the advertisement selected by the user based at least in part on behavior of the user's associated advertisement items and providing the associated advertisement items to the user}, {\color{olive}[...]} .  (\textit{\textbf{Author: PatentGPT}})
& The method of claim 1, further comprising: \MyDBox[orange!30]{monitoring an interaction of the viewing user with the at least one of the plurality of news items}; and \MyDBox[orange!30]{utilizing the interaction to select advertising for display to the viewing user}.

 \\
\bottomrule
\end{tabular}
}
\caption{Examples of three types of plagiarism identified in the texts written by GPT-2 and its training set (more examples are shown in Appendix). Duplicated texts are highlighted in \MyDBox[yellow!30]{yellow}, and words/phrases that contain similar meaning with minimal text overlaps are highlighted in \MyDBox[orange!30]{orange}. {\color{olive}[...]} indicates the texts omitted for brevity. Personally identifiable information (PII) was masked as ***.}
\label{tab:real_examples}
\end{table*}

\section{Introduction}
Language Models (LMs) have become core elements of  Natural Language Processing (NLP) solutions, excelling in a wide range of tasks such as natural language generation (NLG), speech recognition, machine translation, and question answering. 
%\cite{chen2019few, dong2019unified}, speech recognition \cite{park2019specaugment, gulati2020conformer}, machine translation \cite{lample2019cross,yee2019simple}, and question answering \cite{yoon2019pre, chuang2019speechbert}. 
%Among various downstream NLP tasks, especially, 
The development of large-scale text corpora (generally scraped from the Web) has enabled researchers to train increasingly large-scale LMs. Especially, large-scale LMs have demonstrated unprecedented performance on NLG such that LM-generated texts routinely show more novel and interesting stories than human writings do \cite{mccoy2021much}, and the distinction between machine-authored and human-written texts has become non-trivial \cite{uchendu2020authorship,uchendu2021}. 
%The pre-training objective of a LM is often to learn the probability distribution over word sequences. 
%The development of large-scale text corpora (generally scraped from the Web) has allowed researchers to train increasingly large LMs with billions of parameters. 
%which has been a driving force behind recent advances in NLP. 
As a result, there has been a significant increase in the use of LMs in user-facing products and critical applications. %For instance, people can depend on language models to write news articles \cite{wu2019journalism, gagiano2021robustness}, stories \cite{clark2018creative}, poetry \cite{wang2016can} and even song lyrics \cite{sheng2020songmass}. Incorporating a machine in the loop of writing domains (e.g., \cite{clark2018creative, du2022read}) has also gained a lot of attention. 

Concerning the fast-growing adoption of language technologies, it is important to educate citizens and practitioners about the potential ethical, social, and privacy harms of these LMs, as well as strategies and techniques for preventing LMs from adversely impacting people. A body of recent studies has attempted to identify such hazards by examining LMs' capabilities in generating biased and hateful content \cite{pavlopoulos2020toxicity}, spreading misinformation
%through propaganda, scam content, and fake news 
\cite{ahmed2021detecting}, and violating users' privacy \cite{carlini2021extracting}. Particularly, it was shown that machine-generated texts can include individuals' private information such as phone number and email address due to LMs' over-memorization of training samples \cite{carlini2019secret}.

%In particular, our work calls attention to LMs' over-memorization of training samples which tends to result in aforementioned privacy leakage \cite{carlini2019secret, thakkar2021understanding,truex2018towards,arpit2017closer,meehan2020non}. Pivoting back to \citet{carlini2021extracting}, the authors conducted membership inference attacks,\footnote{It is a type of adversarial attacks that aims to predict whether or not a particular example was included in a training set, based on a trained model} and extracted multiple memorized examples containing individuals' personal data such as phone number and email address. They have also found that models' copying behaviors are prone to exacerbate as both the size of LMs and their training data increase, which raises a serious concern about indiscriminately pursuing larger models with larger training corpora. 

%The quality of training corpora is important in model training processes, as learned models reflect the biases present in their training data \cite{bender2021dangers, sheng2020towards}. However, in the case of modern LMs, performing manual review and curation is computationally expensive due to their massive training datasets. In fact, according to \citet{lee2021deduplicating} and \citet{kandpal2022deduplicating}, over-memorization is highly correlated with over-represented training samples. 

Some may argue that, since one's private information %revealing content 
was publicly available in the first place, it is not a problem for LMs to \emph{memorize} and emit it in the generated texts. Still, the current data collection processes (for building training corpora) do not consider how that particular piece of information has been originally released \cite{brown2022does}. For example, it is possible for malicious attackers to hack an individual's private data and intentionally post it online. While training LMs on corpora explicitly intended for public use with creators' consents is ideal, it is challenging to achieve in practice.
%or acquired providers' consent is ideal, but challenging to achieve.   %\citet {brown2022does} claimed that this problem can be mitigated if the model is trained on data explicitly intended for public use (which may require the informed consent of content owners).

Note that over-memorization can be perceived as a threat to the authorship and originality of training instances, as training sets for LMs are routinely downloaded from the Internet without the explicit approval of content owners \cite{brown2022does}. This behavior is known as {\bf plagiarism}--i.e., \emph{the act of exploiting another person’s work or idea without referencing the individual as its author} \cite{ali2011overview}. As shown in Table \ref{tab:real_examples}, for instance, plagiarized content written by a machine may contain not only explicit text overlap but also semantically similar information. Existing memorization studies on LMs have focused only on the memorized sequences that are identical to training sequences \cite{carlini2021extracting, zhang2021counterfactual, lee2021deduplicating}. This motivates our main inquiry of this work: \emph{To what extent (not limited to memorization) do LMs exploit phrases or sentences from their training samples?}

% Another important aspect of LM is the broad usage of fine-tuning paradigm 
On the other hand, the fine-tuning paradigm is widely used in LMs for downstream NLP tasks. Specifically, LMs are initially pre-trained on a massive and diverse corpus and then fine-tuned using a smaller task-specific dataset. This enables LMs to create texts in specific domains such as 
%stories \cite{fan2018hierarchical}, 
poetry \cite{deng2020iterative} and song lyrics \cite{sheng2020songmass}. These tasks require creativity and authenticity, which LMs are prone to fail in. Therefore, the generation outputs of LMs have great moral and ethical implications. Despite increasing efforts to comprehend the over-memorization of pre-trained LMs, to the best of our knowledge, no prior literature has studied on the memorizing behavior of fine-tuned LMs
%which can potentially memorize 
from both pre-training and fine-tuning corpora.

To fill this void of our understanding on the limits of LMs, in this paper, we examine the plagiarizing behaviors of pre-trained and fine-tuned LMs. Our study is guided by two research questions: \textbf{(RQ1) Do pre-trained LMs plagiarize?} and \textbf{(RQ2) Do fine-tuned LMs plagiarize?}.
%\begin{itemize}
%    \item \textbf{(RQ1) Do pre-trained LMs plagiarize?}
%    \item \textbf{(RQ2) Do fine-tuned LMs plagiarize?}
%\end{itemize}
Specifically, we use OpenAI's GPT-2 \cite{radford2019language} for studying these inquiries.\footnote{We chose GPT-2 (instead of more recent LMs such as GPT-3) as it is the latest LM whose replicated training corpus is available. Also, GPT-2 is very popular, ranked as one of the most downloaded LMs from Hugging Face.}
%because  the training corpus of GPT-2 is available.}
%because it allows free usage, and its training data is available online.} 
We first construct a novel pipeline for automated plagiarism detection and use it to 
% Our pipeline is designed to enhance paraphrase detection accuracy of \citet{sanchez2015adaptive}.
identify three types of plagiarism (i.e., \emph{verbatim}, \emph{paraphrase}, \emph{idea} plagiarism)  from passages generated by pre-trained GPT-2 with different combinations of model sizes and decoding methods. For RQ2, three GPT-2 models are fine-tuned using datasets in scholarly writing and legal domains, which are later used for comparing plagiarism from pre-training and fine-tuning corpora. 

%GPT-2 is a transformer-based LM that comes in 4 different sizes — small, medium, large, and xl, with 124M, 355M, 774M, and 1.5B parameters, respectively

% results
Our results demonstrate that \ul{machine-generated texts do plagiarize from training samples, across all three types of plagiarism}. We discover three attributes that impact LMs' plagiarism: 1) \emph{model size}: larger models plagiarize more from a training set than smaller models; 2) \emph{decoding methods:} decoding the outputs after limiting the output space via top-\emph{p} and top-\emph{k} strategies are positively related to heightened plagiarism levels as opposed to a raw vocabulary distribution; 3) \emph{corpus similarity and homogeneity:} a higher corpus similarity level across pre-training and fine-tuning corpora, as well as within fine-tuning corpora, enhances the degree of plagiarism for a fine-tuned model.

%Contributions of our paper are summarized as follows:
In summary, our work makes the following contributions:
\begin{itemize}[leftmargin=\dimexpr\parindent+0.1\labelwidth\relax, noitemsep]

\item By leveraging a BERT-based classifier together with Named Entity Recognition (NER) on top of \citet{sanchez2015adaptive}'s plagiarism detection model, we empirically highlight that LMs do more than copying and pasting texts in a training set; it further rephrases sentences or mimics ideas from other writings without properly crediting the source. %consent. %\thai{Why we need this? What are the potential challenges? Should be briefly addressed in previous paragraphs.}.

\item To the best of our knowledge, this is the first work to systematically study the plagiarizing behavior of \textit{fine-tuned} LMs. %Specifically, we find that the intra- and inter-corpus similarity would significantly influence the fine-tuned LMs' plagiarizing behavior. 
Specifically, we find that restricting intra- and inter-corpus similarity can considerably decrease the rate of plagiarism.

\item We provide a deeper understanding of the factors that influence LMs' plagiarizing patterns such as model size, decoding strategies, and a fine-tuning corpus. Our results add value to the ongoing discussion around memorization in modern LMs and pave the way for future research into designing robust, reliable, and responsible LMs.
\end{itemize}

%\todo{probably need to rephrase the contribution part? maybe (1) improve pipeline and use it to study LM's plagiarism pattern (2) study fine-tune models' plagiarism which no one has done before (3) provide more understandings to shed lights on designing better LMs}

\section{Related Work} 
%\thai{Related works are to position our paper with other previous papers. I recommend removing non-relevant details such as how to mitigate over-memorization, and focus on describing ``how our paper improves/novel/different/innovative/solve from problems compared to previous works" (while keeping the same citations).}

\subsection{Memorization in LMs} 
There is a growing body of literature that aims to study the memorization of neural LMs by recovering texts in the training corpus \cite{salem2020updates,leino2020stolen} or extracting artificially injected canaries \cite{, mireshghallah2021privacy,zanella2020analyzing}. \citet{carlini2021extracting} and \citet{brown2022does} emphasized that data memorization can intentionally or unintentionally lead to sensitive information leakage from a model’s training set. Meanwhile, recent studies \cite{lee2021deduplicating, kandpal2022deduplicating} have shown that training data of LMs tend to contain a large number of near-duplicates, and overlapping phrases included in near-duplicates significantly account for memorized text sequences. In order to distinguish rare but memorized texts from trivial examples, \citet{zhang2021counterfactual} presented a notion of counterfactual memorization which measures a difference in the expected performance of two models trained with or without a particular training sample. 

Still, none of these works have explored beyond text overlap. The most relevant research to ours is \citet{mccoy2021much}, which analyzed the novelty of machine-generated texts. Although authors found 1,000 word-long duplicated passages from a training set, they concluded that neural LMs can integrate familiar parts into novel content, rather than simply copying training samples. However, because they did not directly compare identified novel content with training samples, the level of plagiarism is uncertain.

\subsection{Automatic Plagiarism Detection}
Automated extrinsic plagiarism detection, in general, can be divided into two subtasks: document retrieval and text alignment. While document retrieval focuses on fetching all documents that potentially have plagiarized an existing document, the text alignment subtask detects the location and content of plagiarized texts. \citet{alzahrani2015arabic} retrieved candidate documents that share exactly copied sequences and computed the similarity between overlapping 8-grams. There are diverse ways to measure text similarity with segmented document pairs. For example, \citet{kuppers2012set} calculated the Dice coefficient between 250 character chunks of passage pairs, and \citet{shrestha2013using} implemented the Jaccard similarity with n-grams.

More recently, there has been continuous efforts in incorporating word embedding and advanced machine learning or deep learning models for plagiarism detection. \citet{agarwal2018deep} used Convolutional Neural Network (CNN) to obtain the local region information from n-grams and applied Recurrent Neural Network (RNN) to capture the long-term dependency information. Similarly, \citet{altheneyan2020automatic} viewed the task as a classification problem and developed a support vector machine (SVM) classifier using several lexical, syntactic, and semantic features. In our proposed method, we combine conventional similarity measurements and state-of-the-art models to maximize the detection performance.

%\citet{gharavi2016deep} extracted word vectors using the word2vec algorithm and applied two similarity metrics: Cosine similarity, and Jaccard similarity. Instead of using well-established similarity scores bounded by particular thresholds, \citet{altheneyan2020automatic} has viewed the task as a classification problem and developed a support vector machine (SVM) classifier using several lexical, syntactic, and semantic features. Specifically for paraphrase detection, \citet{agarwal2018deep} relied on Convolutional Neural Network (CNN) to obtain the local region information from n-grams and Recurrent Neural Network (RNN) to capture the long-term dependency information.

\section{Plagiarism: Definition and Detection}

%\lee{First, answer "what is plagiarism?" using representative literature first.}

\subsection{Taxonomy of Plagiarism}
Plagiarism occurs when any content including text, source code, or audio-visual content is reused without permission or citation from an author of the original work \cite{park2003other,cosma2008towards}. It has been a longstanding problem, especially in educational and research institutions or publishers, given the availability of digital artifacts \cite{clarke2006plagiarism}. Plagiarism can severely damage academic integrity and even hurt individuals' reputation and morality \cite{east2010judging}. To detect such activities, it is necessary to have extensive knowledge about plagiarism forms and classes. 

In this work, we focus on the three most commonly studied plagiarism types: \emph{verbatim} plagiarism, \emph{paraphrase} plagiarism, and \emph{idea} plagiarism. Verbatim plagiarism, which can be considered as the most naive approach, is to directly copy segments of others' documents and paste them into their writings \cite{dhammi2016plagiarism}. To make plagiarism less obvious, one may incorporate paraphrase plagiarism by replacing original words with synonyms or rearrange word orders \cite{barron2013plagiarism}. Similarly, back translation, using two independent translators to translate sentences back and forth, is common in generating paraphrases. Lastly, reuse of the core idea from the original content, also known as idea plagiarism, is a challenging case for an automatic detection due to limited lexical and syntactic similarities. Hence, existing literature (e.g., \citet{vani2017detection, gupta2016plagiarism}) specified the task to capture whether a document embeds a summary of another document. While paraphrase plagiarism targets sentence-to-sentence transformations, idea plagiarism reads a chunk of the content and condenses its main information into fewer sentences (or vice versa). In essence, in this work, we adopt the following definition of three plagiarism types:
\begin{itemize}[leftmargin=\dimexpr\parindent+0.1\labelwidth\relax]
   \item \textbf{Verbatim plagiarism}: exact copies of words or phrases without transformation.
    \item \textbf{Paraphrase plagiarism}: synonymous substitution, word reordering, and/or back translation.
    \item \textbf{Idea plagiarism}: representation of core content in an elongated form.
\end{itemize}

%\lee{any citations to back up these definitions?}

%\begin{figure}[h]
%  \lee{need to improve this figure as it's not clear}
%  \centering
%  \includegraphics[width=\linewidth]{image/pipeline.png}
%\caption{Plagiarism Detection Pipeline}
%\label{fig:pipeline}
%\end{figure}

\subsection{Automatic Detection of Plagiarism}
%\thai{This section can be significantly improved with a diagram.}
\label{sec:plagiarism_framework}

%It is computationally expensive to iteratively compare all document pairs to detect plagiarism within a large text corpus. To improve a runtime, 
In this section, we introduce a two-step approach for automated plagiarism detection. Suppose we have $n$ documents in a corpus $D$=\{$d_1$, $d_2$, ... $d_n$\} and a query document $d_q$. The goal is to identify a pair of ``plagiarized" text segments ($s_1$, $s_2$) such that $s_1$ (resp. $s_2$) is a text segment within a document $d_i \in D$ (resp. $d_q$).

\vspace{0.1in}
\noindent
{\bf Step 1 (Finding Top-$n'$ Candidate Documents)}: First, for the given query document $d_q$, we aim to quickly narrow down to top-$n'$ documents (out of $n$ documents, where $n' \ll n$) which are likely to contain plagiarized pieces of texts.
%
%Below we describe the processes of automated plagiarism type identification. %Figure \ref{fig:pipeline} shows a visual representation of our pipeline. 
%We first store a training corpus to the search engine in order to effectively fetch relevant document pairs. We then apply text alignment to those pairs and detect plagiarism types.  
%
%\subsubsection{Candidate Document Retrieval}
\label{sec:document}
%explain elastic search
%The first step of our approach is to distinguish a list of candidate documents from training samples that are most relevant to machine-generated documents. Here 
To do this, we utilize a document similarity score as a proxy for plagiarism. Since recent LMs are generally trained on gigantic corpora, it is non-trivial to store them locally and compute a pair-wise document similarity. Hence, we implement a search engine using Elasticsearch\footnote{\url{https://www.elastic.co/elasticsearch/}}, an open-source search engine built on Apache Lucene that provides a distributed RESTful search service with a fast response time. 
After storing the entire training documents $D$ in Elasticsearch, using a machine-generated document as the query document $d_q$, we retrieve top-$n'$ most-similar documents.
%initiate the search process by setting the whole content of the original document (in our case, machine-generated document) as queries. We clean queries by removing stopwords and lemmatizing. It then automatically calculates similarities between stored documents and inserted queries and fetches top \emph{N} candidate documents that acquire the highest similarity scores. 
Elasticsearch utilizes the Okapi-BM25 algorithm \cite{robertson1995okapi}, a popular bag-of-words ranking function, by default. We used $n'=10$ in experiments for the sake of time efficiency.\footnote{We performed a post-hoc analysis with a smaller ($n'=5$) and a larger value ($n'=30$) of $n'$ using GPT-2 xl to gauge its potential effects on identified plagiarism rates. The results showed a marginal difference (e.g., 1.46\% ($n'=5$) vs. 1.54\% ($n'=30$) for temperature setting), indicating that the choice of the $n'$ value does not drastically influence our findings.}

\vspace{0.1in}
\noindent
{\bf Step 2 (Finding Plagiarized Text Pairs and Plagiarism Type)}:
Next, using the identified $n'$ candidates \{$d_1$, $d_2$, ..., $d_{n'}$\} for the query document $d_q$, we aim to find plagiarized text pairs ($s_1$, $s_2$) such that $s_2$ is one of three types of plagiarism against $s_1$.
%
%\subsubsection{Plagiarism Type Identification}
%explain PAN2015 approach and how we improve its performance
%\textbf{Baseline.} 
%
For this task, we exploit text alignment algorithms that locate and extract most-similar contiguous text sequences between two given documents. Such text alignment algorithms are applicable to various tasks such as 
%information retrieval \cite{davis1997free, semmar2007arabic}, 
text-reuse detection \cite{sochenkov2016exactus} and translation alignment \cite{lin2020pre}. 
In particular, we employ the improved version of the winning method at the plagiarism detection competition of PAN 2014.\footnote{\url{https://pan.webis.de/clef14/pan14-web/text-alignment.html}} Following, we explain details on \citet{sanchez2015adaptive} and our improvement strategies.

%Further details on \citet{sanchez2015adaptive} are elaborated in Section \ref{sec:appendix1}.

\vspace{0.1in}
\noindent \textbf{Current Approach (\citet{sanchez2015adaptive}).} Their methods consist of five steps which include (1) text-preprocessing (lower-casing all characters, tokenizing, and stemming); (2) obfuscation type identification (verbatim/random/translation/summary obfuscation); (3) seeding (deconstructing long passages into smaller segments and finding candidate pairs through sentence-level similarity measurement given two documents); (4) extension (forming larger text fragments that are similar via clustering); and (5) filtering (removing overlapping and short plagiarized fragments). In summary, they transform the suspicious and source sentences as term frequency–inverse document frequency vector weights and then calculate the similarity between the sentence pairs using the dice coefficient and cosine measure. Adaptive parameter selection is achieved by testing two settings recursively for the summary obfuscation corpus and the other three corpora. 
%For the purpose of our study, random and translation obfuscation types are grouped as paraphrase plagiarism, and summary obfuscation is considered as idea plagiarism.

% \usepackage{booktabs}
% \usepackage{multirow}
\begin{table}[]{
\centering
\small
\hfill{}
\begin{tabular}{@{}cccclccc@{}}
\toprule
\multirow{2}{*}{Scores} & \multicolumn{3}{c}{PanDataset}  &  & \multicolumn{3}{c}{GptPlagiarismDataset} \\ \cmidrule(lr){2-4} \cmidrule(l){6-8} 
                        & Verbatim & Paraphrase & Idea &  & Verbatim    & Paraphrase    & Idea    \\ \midrule
Precision               & 0.995    & 1.00       & 1.00    &  & 0.96        & 0.846         & 0.99       \\
Recall                  & 0.986    & 0.723      & 0.412   &  & 0.87        & 0.785         & 0.3        \\ \bottomrule
\end{tabular}
\hfill{}
}
\caption{Evaluation results of our plagiarism detection pipeline. For PanDataset, we perform the evaluation in a binary classification setting (e.g., verbatim plagiarism vs. no plagiarism). Since GptPlagiarismDataset does not take into account document pairs without plagiarism, we adopt a multi-nomial classification setting (e.g., verbatim plagiarism vs. paraphrase/idea plagiarism).}
\label{tab:eval_pipeline}
\end{table}

%Among several works dedicated to plagiarism detection--i.e., Step 2, \citet{sanchez2015adaptive} is the only one that publicly shared a source code with a reported robust performance that we can utilize, particularly for verbatim and paraphrase detection. Moreover, their approach enables us to quickly identify the \emph{longest} plagiarized substrings unlike existing plagiarism detectors trained and evaluated on labeled sentence pairs \cite{shahmohammadi2021paraphrase, socher2011dynamic}. 
\vspace{0.1in}
\noindent \textbf{Our Improvements.} To verify the effectiveness of \citet{sanchez2015adaptive} on our corpus, we manually inspected 200 plagiarism detection results. For a fair comparison, the number of sentence pairs in each category (none/verbatim/paraphrase/idea plagiarism) was equally distributed. Our evaluation revealed that \citet{sanchez2015adaptive} induces more false positives than their reported performance, specifically in detecting the paraphrase type plagiarism (0.51 in precision%\thai{why we need i.e., here? can we remove it?}
). It resulted from the model's tendency of labeling near-duplicates with one character difference as paraphrases (should be the ``verbatim" plagiarism type) and its inability to distinguish a minor entity-level discrepancy such as numerical values or dates. To minimize such errors,  after \citet{sanchez2015adaptive} retrieves all paraphrased text segments, we post-process segments by chunking them into sentences with NLTK\footnote{\url{https://www.nltk.org}}'s sentence tokenizer and apply a RoBERTa-based paraphrase identification model \cite{morris2020textattack}\footnote{The RoBERTa classifier has achieved 91.17\% accuracy on the evaluation set from the MSRP corpus (\url{https://www.microsoft.com/en-us/download/details.aspx?id=52398}).} and Named-Entity Recognition (NER)\footnote{We use SpaCy library (\url{https://spacy.io}).} as additional validators. Specifically, when there is at least one sentence pair whose probability score  (from the paraphrase detection model) ranges from 0.5 to 0.99\footnote{We specified 0.99 as the upper bound to avoid near-duplicate pairs.} and have the exactly matching set of entities, we ultimately accept the plagiarism result by \citet{sanchez2015adaptive}. This additional restriction resulted in the following precision scores: 0.92 for no plagiarism, 1.0 for verbatim type, 0.88 for paraphrase type, and 0.62 for idea type. To gauge both precision and recall, we utilize two additional labeled datasets, PanDataset and GptPlagiarismDataset (refer to Appendix \ref{sec:appendix7} for more details on datasets). Both precision and recall scores of each label are reported in Table \ref{tab:eval_pipeline}.
Note that at the end, \ul{our plagiarism detection pipeline has high precisions at the cost of low recalls, implying that the number of plagiarism cases we report subsequently is only a ``lower-bound" estimate of plagiarism rates that actually exist}.
For subsequent analyses, we utilize two hyperparameters: (1) the minimum character count of common substrings between the two documents for verbatim plagiarism is set to 256; (2) the minimum character count permitted on either side of a plagiarism case is set to 150. These thresholds are much  stricter than minimum 50 tokens (i.e., on average 127 characters) employed by existing works \cite{lee2021deduplicating, carlini2022quantifying}. Again, \ul{this ensures that our following report on RQ1 and RQ2 is the ``lower-bound" estimate of plagiarism frequencies}.

%We employ this improved version of the tool for the subsequent experiments. To loosen the definition of plagiarism, we convert two hyperparameter values: (1) the minimum character count of common substrings between the two documents for verbatim plagiarism is reduced from 256 to 50; (2) the minimum character count permitted on either side of a plagiarism case is reduced from 150 to 50.

\section{RQ1: Do Pre-trained LMs Plagiarize?}

%\thai{The word ``GPT-2" was not mentioned in the RQ1 introduction. You can remove ``GPT-2" here, then explain why you use ``GPT-2" in the next paragraph, and how ``GPT-2" represents similar LMs}

\subsection{Experimental Setup}
%\thai{You might want to move unimportant footnotes to Appendix and mention here that all details are in the Appendix, to leave space for other improvements if needed.}
\textbf{Dataset.} GPT-2 is pre-trained on WebText, containing over 8 million documents retrieved from 45 million Reddit links. %After data de-duplication and some heuristic-based cleaning, its reported final size is over 8 million documents for a total of 40 GB of text.
Since OpenAI has not publicly released WebText, we use OpenWebText which is an open-source recreation of the WebText corpus.\footnote{\url{https://skylion007.github.io/OpenWebTextCorpus/}} It has been reliably used by prior literature \cite{kandpal2022deduplicating,liu2019roberta}.

%Given that the size of OpenWebtext corpus matches the size described in \citet{radford2019language}, and previous literature \cite{radford2019language,liu2019roberta} already used it, we assume it is a reliable source.
%\thai{You can cite other similar papers that make the same assumptions.}.

%\\\vspace{0.5em}
 \vspace{0.05in}\noindent \textbf{Model.} GPT-2 is an auto-regressive language model predicting one token at a time in a left-to-right fashion. That is, the probability distribution of a word sequence can be calculated through the product of conditional next word distributions. In response to an arbitrary prompt, GPT-2 can adapt to its style and content and generate artificial texts. GPT-2 comes in 4 different sizes — small, medium, large, and xl, with 124M, 355M, 774M, and 1.5B parameters, respectively. We utilize all of them for analyses.

%According to \citet{radford2019language}, the smallest model is equivalent to the original GPT \cite{radford2018improving}, and the second smallest is same as the largest model from BERT \cite{devlin2018bert}.

%\\\vspace{0.5em}
 \vspace{0.05in}\noindent \textbf{Text Generation.} 
Given that GPT-2 relies on the probability distribution when generating word-tokens, there exist various decoding methods which are well known to be critical for performance in text generation \cite{ippolito2019human}. We primarily consider the following decoding algorithms:  

\begin{itemize}[leftmargin=\dimexpr\parindent+0.1\labelwidth\relax]
    \item Temperature \cite{ackley1985learning}: control the randomness of predictions by dividing the logits by \emph{t} before applying softmax
    \item Top-\emph{k} \cite{fan2018hierarchical}: filter the \emph{k} most likely next words and redistribute the probability mass 
    \item Top-\emph{p} \cite{holtzman2019curious}: choose from the smallest possible set of words whose cumulative probability exceeds the probability \emph{p}
\end{itemize}

It is reported that increasing parameter values (\emph{t, k, p}) can notably improve the novelty of machine-generated texts but may also deteriorate their quality sides \cite{mccoy2021much}. Conversely, smaller parameter values tend to yield dull and repetitive sentences \cite{holtzman2019curious}. 

Considering the difficulties in hyper-parameter tuning that can confidently guarantee high-quality machine-authored texts, we use off-the-shelf GPT-2 Output Dataset\footnote{\url{https://github.com/openai/gpt-2-output-dataset}} provided by OpenAI. This dataset has been reliably used by \citet{kushnareva2021artificial} and \citet{wolff2020attacking} for neural text detection. Specifically, It contains 250,000 texts generated by four versions of the GPT-2 model with aforementioned decoding approaches. Owners of the repository have informed us that they used a `<|endoftext|>' token as a prompt and set \emph{t}=1, \emph{k}=40, $0.8{<}\emph{p}{<}1$.\footnote{Equivalent to existing literature \cite{mccoy2021much, delucia2020decoding}, we only report results of these specific hyperparameters because they were recommended by GPT-2 creators \cite{radford2019language} Also, our findings on the decoding methods were validated by additional experiments with more diverse parameter values.}. In total, there are 12 (i.e., 4 model size * 3 decoding methods) combinations, and we analyze 10,000 documents in each combination.

%\cnote{the legend is covering the data, maybe move it to lower right?}
\begin{figure}[tb!]
  \centering
  \includegraphics[width=0.95\linewidth]{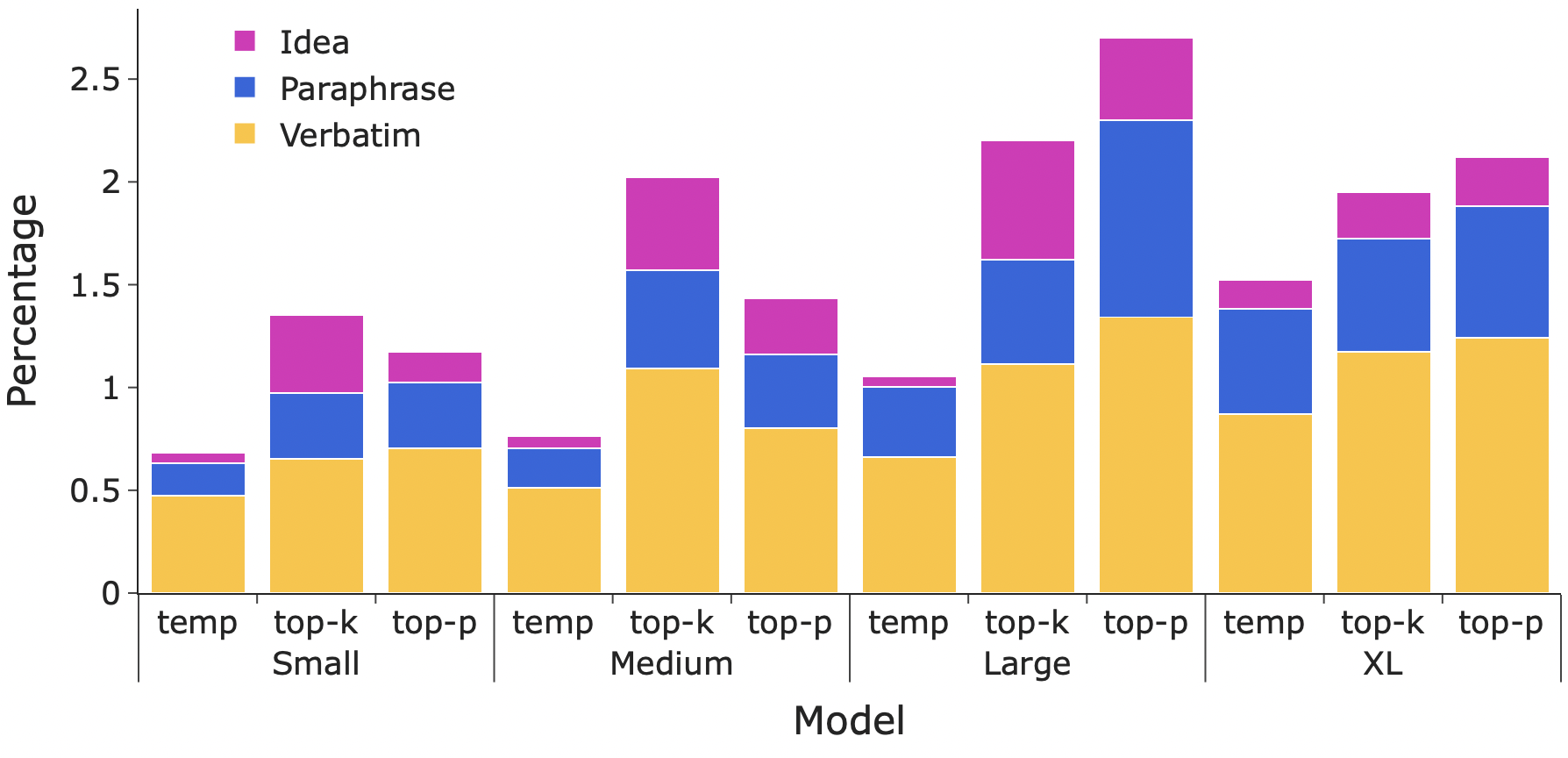}
    \caption{Document percentage w.r.t. three plagiarism types from pre-training data}
    %\lee{redraw this graph--cannot read legends or labels of axis}}
  \label{fig:gpt2_plagiarism}
%   \vspace{-0.5cm}
\end{figure}
% \vspace{-2mm}

\subsection{Results}
We discover that \ul{pre-trained GPT-2 families do plagiarize from the OpenWebText}. Figure \ref{fig:gpt2_plagiarism} illustrates the percentage of unique machine-written documents regarding three plagiarism types based on different model sizes and decoding strategies\footnote{Please note that sentences with proper quotation marks within identified plagiarism cases were excluded from the analyses, as they do not constitute plagiarism.}. Consistent with \cite{carlini2021extracting, levy2021investigating}, the larger the model size became, the higher occurrences of plagiarism were observed when using temperature sampling. %\todo{do these works only consider temp setting? or they also consider top-k and top-p? It seems a bit strange only mentioned the temp setting is consistent. Also do you introduce the difference of top-k/temp/top-p?} 
%We also find that, in addition to verbatim plagiarism (which is equivalent to memorized substrings), the other two types of plagiarism surged alongside the model size. 
The general trend still holds when GPT-2's word token is sampled with top-\emph{k} and top-\emph{p} truncation except for the xl model size. 
% However, the observed trend does not hold when GPT-2's word token is sampled with top-\emph{k} and top-\emph{p} truncation: 
However, interestingly, plagiarism frequencies were the highest when GPT-2 large models were used, not xl.  %\thai{Do you have a good guess why this happens?}
%\todo{any explanation why xl got slightly better than large? otherwise the larger, the more plagiarization conclusion is a bit problematic?}
%
We also find that decoding methods affect models' plagiarism. More precisely, top-\emph{k} and top-\emph{p} sampling are more strongly associated with plagiarism than decoding with temperature regardless of the model size. We conjecture that this discrepancy is due to the fact that top-\emph{k} and top-\emph{p} decoding methods disregard less probable tokens unlike random sampling, which may push models to choose a memorized one as a next token. 

%To sum up, these observations bring to light the dangers of mindlessly pursuing models with more parameters and complicated decoding approaches. 
%\thai{Maybe you can briefly mention what is the implications/applications from this last observation, and say we will analyze in detail later in Section 7.} 
 
% \usepackage{booktabs}
% \usepackage{multirow}
\begin{table*}[]
\centering
{\small
\hfill{}
\begin{tabular}{@{}ccccclccc@{}}
\toprule
\multirow{2}{*}{Model}                                                       & \multirow{2}{*}{Decoding} & \multicolumn{3}{c}{Plagiarism from Pre-Training Data}                       &  & \multicolumn{3}{c}{Plagiarism from Fine-Tuning Data}                                 \\ \cmidrule(lr){3-5} \cmidrule(l){7-9} 
                                                                             &                           & Verbatim    & Paraphrase            & Idea                  &  & Verbatim             & Paraphrase            & Idea                  \\ \midrule

\multirow{3}{*}{\begin{tabular}[c]{@{}c@{}} \textcolor[HTML]{0000CD}{\textbf{Pre-trained}}\\ \textcolor[HTML]{0000CD}{\textbf{GPT}}\end{tabular}}   & temp                      & 47 (0.47\%) & 16 (0.16\%)           & 5 (0.05\%)            &  & \multicolumn{3}{c}{\multirow{3}{*}{N/A}}                             \\
                                                                             & top-\emph{k}                     & 65 (0.65\%) & 32 (0.32\%)           & 38 (0.38\%)           &  & \multicolumn{3}{c}{}                                                 \\
                                                                             & top-\emph{p}                     & \textbf{70 (0.7\%)}  & 32 (0.32\%)           & 15 (0.15\%)           &  & \multicolumn{3}{c}{}                                                 \\
                                                                        \midrule 
                                                                        \midrule

\multirow{3}{*}{\begin{tabular}[c]{@{}c@{}}\textcolor[HTML]{FF1493}{\textbf{Patent}}\\ \textcolor[HTML]{FF1493}{\textbf{GPT}}\end{tabular}}        & temp                      & 0 (0\%)     & 36 (0.36\%)           & 21 (0.21\%)            &  & 0 (0\%)              & 32 (0.32\%)           & 17 (0.17\%)           \\
                                                                             & top-\emph{k}                     & 0 (0\%)     & \textbf{171 (1.71\%)} & \textbf{161 (1.61\%)} &  & 0 (0\%)              & 2 (0.02\%)             & 0 (0\%)               \\
                                                                             & top-\emph{p}                     & 0 (0\%)     & 94 (0.94\%)           & 130 (1.3\%)           &  & 0 (0\%)              & 3 (0.03\%)            & 0 (0\%)               \\ \midrule
\multirow{3}{*}{\begin{tabular}[c]{@{}c@{}}\textcolor[HTML]{FF1493}{\textbf{Cord19}}\\ \textcolor[HTML]{FF1493}{\textbf{GPT}}\end{tabular}}        & temp                      & 0 (0\%)     & 6 (0.06\%)            & 6 (0.06\%)            &  & 43 (0.43\%)          & 90 (0.9\%)           & 42 (0.42\%)           \\
                                                                             & top-\emph{k}                     & 0 (0\%)     & 79 (0.79\%)           & 122 (1.22\%)          &  & 46 (0.46\%)          & \textbf{548 (5.48\%)} & \textbf{485 (4.85\%)} \\
                                                                             & top-\emph{p}                     & 2 (0.02\%)  & 57 (0.57\%)           & 79 (0.79\%)           &  & \textbf{72 (0.72\%)} & 388 (3.88\%)          & 228 (2.28\%)          \\ \midrule
\multirow{3}{*}{\begin{tabular}[c]{@{}c@{}}\textcolor[HTML]{FF1493}{\textbf{ArxivAbstract}}\\ \textcolor[HTML]{FF1493}{\textbf{GPT}}\end{tabular}} & temp                      & 0 (0\%)     & 0 (0\%)               & 0 (0\%)               &  & 0 (0\%)              & 3 (0.03\%)            & 0 (0\%)               \\
                                                                             & top-\emph{k}                     & 0 (0\%)     & 0 (0\%)               & 1 (0.01\%)            &  & 0 (0\%)              & 0 (0\%)               & 0 (0\%)               \\
                                                                             & top-\emph{p}                     & 0 (0\%)     & 2 (0.02\%)            & 0 (0\%)               &  & 0 (0\%)              & 2 (0.02\%)            & 0 (0\%)               \\ 
\bottomrule
\end{tabular}
\hfill
}
\caption{Number (\%) of machine-written documents w.r.t. three plagiarism types from pre-training \& fine-tuning data. \textcolor[HTML]{0000CD}{Blue} represents the pre-trained model, whereas \textcolor[HTML]{FF1493}{pink} represents the fine-trained model. in A total number of documents we generated for each model and decoding methods is 10,000.}
\label{tab:finetuned_plagiarism}
\end{table*}

\subsection{Qualitative Examination of Plagiarized Texts}

% \usepackage{booktabs}

%\begin{table}[]
%\begin{tabular}{@{}cccc@{}}
%\toprule
%\multicolumn{1}{l}{Model Size} & \multicolumn{1}{l}{Verbatim} & \multicolumn{1}{l}{Paraphrase} & \multicolumn{1}{l}{Idea} \\ \midrule
%Small                          & 18.1                        & 19.2                          & 42.4                    \\
%Medium                         & 17.1                        & 27.3                          & 54.3                    \\
%Large                          & 18.5                        & 23.1                          & 42.5                    \\
%xl                             & 17.0                         & 26.4                          & 47.2                    \\ \bottomrule
%\end{tabular}
%\caption{Proportion (\%) of identified plagiarized sentences (by word count) within documents. The values are computed by taking a mean of median values of three decoding methods.}
%\label{tab:word_proportion}
%\end{table}

\begin{figure}[tb!]
  \centering
  \includegraphics[width=0.9\linewidth]{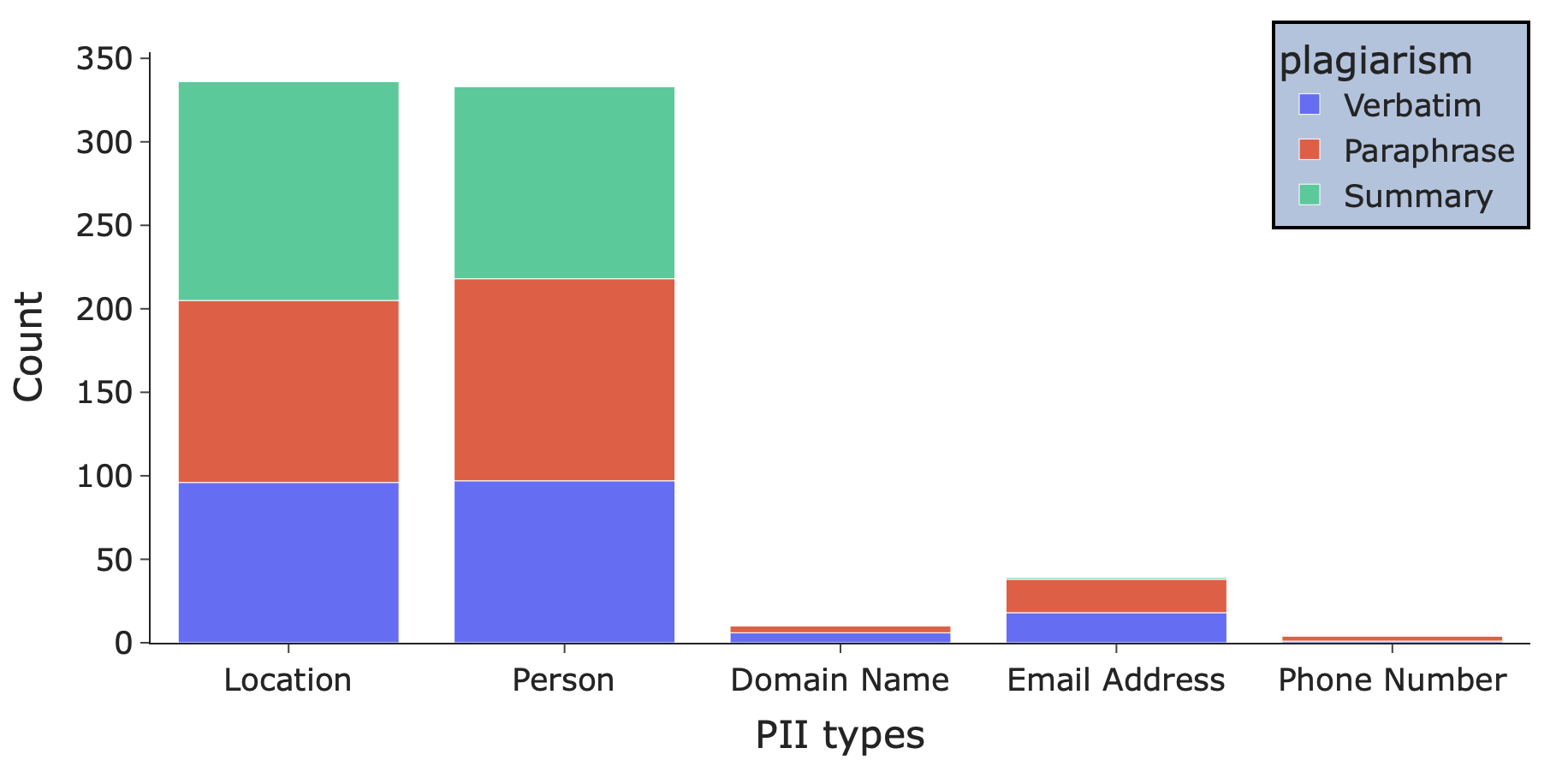}
    \caption{Number of unique PII-exposing substrings associated with plagiarism categories}
  \label{fig:pii_count}
\end{figure}

\textbf{Lengths and Occurrences.}
%Table \ref{tab:plagiarism_examples} presents several plagiarized content we discovered for each category. 
Motivated by prior memorization studies \cite{carlini2022quantifying, lee2021deduplicating}, we inspect lengths and occurrences of texts that are associated with verbatim plagiarism. We find that the median length of memorized texts is 483 characters, and the longest texts contain 5,920 characters. In order to efficiently count the occurrences of plagiarized strings within OpenWebText, we utilize the established Elasticsearch pipeline, which includes setting plagiarized texts as search queries and retrieving documents that embed provided texts.\footnote{By default, Elasticsearch does not allow searches to return more than the top 10,000 matching hits.} 
We find that some memorized sequences are from highly duplicated texts throughout the training corpus: the newsletter sign-up text \footnote{``newsletter sign up continue reading the main story please verify you're not a robot by clicking the box. invalid email address. please re-enter...''} appeared at most 9,978 times and was memorized. Still, there exist many instances where models memorize without seeing them more than two times. While the median of occurrences for memorized texts is 6, sequences related to paraphrase or idea plagiarism are prone to not appear at all from training samples (median = 0).

\vspace{0.05in}
\noindent \textbf{Inclusion of Sensitive Information.} We now turn our attention to whether sequences associated with three plagiarism types contain individuals' personal or sensitive data. To achieve this, we use Microsoft's Presidio analyzer,\footnote{\url{https://microsoft.github.io/presidio/analyzer/}} a Python toolkit for personally identifiable information (PII) entity detection (e.g., credit card information, email address, phone number). There are a total of 1,193 unique text sequences (verbatim: 388, paraphrase: 507, and idea: 298) plagiarized by pre-trained GPT-2. We set a confidence threshold to 0.7. A total number of plagiarized documents that reveal PII entities is shown in Figure \ref{fig:pii_count}. 
Of 1,193 plagiarized sequences, nearly 28\% include at least one element of location information and a person's full name. Although none of highly sensitive information (e.g., driver license number, credit card information, bank number, social security number, and IP address) is revealed, the results show a possibility of machine-generated texts disseminating personal data such as phone number and email address through all three types of plagiarism.
%verbatim plagiarism but also through paraphrase and idea plagiarism.
 
\section{RQ2: Do Fine-tuned LMs Plagiarize?}

\subsection{Experimental Setup}
\textbf{Dataset.} We choose public English  datasets related to scholarly and legal writings because plagiarism is deemed more sensitive and intolerable in these domains \cite{pecorari2008academic}. Three datasets are:
\begin{itemize}[leftmargin=\dimexpr\parindent+0.1\labelwidth\relax, noitemsep]
 \item \textbf{ArxivAbstract}: includes 250,000 randomly selected abstracts on arxiv.org, from the start of the site in 1993 to the end of 2019 \cite{r_stuart_geiger_2019_2533436}. It covers a wide range of disciplines (e.g., Physics, Computer Science, Economics).
 \item \textbf{Cord-19}: consists of 500,000 scholarly articles about the COVID-19 virus \cite{wang2020cord}. Medicine (55\%), Biology (31\%), and Chemistry (3\%) are primary domains of this corpus. 
 For fine-tuning purposes, we randomly sample 200,000 documents.\footnote{Since most articles in CordD-19 exceed the length of 1,024 tokens, we only consider the first five paragraphs starting from the `Introduction' section.}
 \item \textbf{PatentClaim}: is provided by \citet{lee2020patent} and has 277,947 patent claims in total.
\end{itemize}

%\\\vspace{0.5em}
\noindent \textbf{Model.} Using these datasets, we fine-tune three independent GPT-2 small models\footnote{Due to constraints of computing resource, we only fine-tune the GPT-2 small variation.} and denote them as \emph{ArXivAbstractGPT}, \emph{Cord19GPT}, and \emph{PatentGPT}, respectively. The details on training configurations can be found in Appendix \ref{sec:appendix5}.

%\\\vspace{0.5em}
 \vspace{0.05in}\noindent \textbf{Text Generation.} For three fine-tuned models, we manually create 10,000 machine-generated texts using the same prompt and parameter settings as GPT-2 Output Dataset.

\subsection{Results} 
%\thai{This results are not interesting to the readers. I saw that in Table 3, some number shows that fine-tuned model plagiarizes on the pre-training data more than pre-trained models. This is more interesting because often the more we train on different data the less plagiarism. Then, you build some hypothesis that can make this observation happens, which motivates the next Section 6.}
%\cnote{Not sure how EMNLP paper usually do, but I think most ML papers put titles below table/figure. Better to check it.}

%\cnote{It's a bit hard to get the information from Fig 2/Fig 3 as you put too many different settings together. I would suggest to control the variables for each figure, i.e., only compare different model size or decoding methods in one figure and list the other comparison in another figure. Same applies to Fig 4/5}

%\begin{figure}[h]
%  \centering
%  \includegraphics[width=\linewidth]{image/finetuned_openwebtext_examples.png}
%    \caption{Distribution of Plagiarism Categories w.r.t. Model and Decoding Methods (Against Pre-training Data)}
%  \label{fig:finetuned_webtext_plagiarism}
%\end{figure}

We compare plagiarizing behaviors of three fine-tuned models using both pre-training (OpenWebText) and fine-tuning datasets (PatentClaim, Cord-19, ArxivAbstract) in Table \ref{tab:finetuned_plagiarism}. Our findings show that \ul{fine-tuning significantly reduces verbatim plagiarism cases from OpenWebText}. This observation aligns with GPT-2's outstanding adaptability to the writing styles of a new corpus. \ul{Yet, not all fine-tuned models are plagiarism-free}; for PatentGPT and Cord19GPT, the remaining plagiarism types regarding OpenWebText occurred more frequently than the pre-trained GPT. Meanwhile, ArxivAbstractGPT barely plagiarized texts from OpenWebText. Interestingly, models' plagiarism behaviors change when we compare their generated texts against the fine-tuning samples. Cord19GPT was strongly affiliated with plagiarism, whereas the other two models were not. 

These results suggest that, although three models are fine-tuned in a similar setting (regarding dataset size and training duration), their patterns of plagiarism vary. We hypothesize that there are external factors that affect models' plagiarism. For example, if fine-tuning and pre-training corpora have multiple similar or duplicated content, the fine-tuned model would have been immensely exposed to it and may have started to remember it. \citet{lee2021deduplicating} has shown a positive relationship between memorized sequences and their frequencies in a training set. Similarly, it is also possible that over-exposure to particular texts may have been resulted from similar documents within fine-tuning data. Next, we analyze a corpus similarity between fine-tuning data and pre-training data and a homogeneity of fine-tuning data in Section \ref{sec:corpus_sim} to verify our hypotheses.

\section{Plagiarism v.s. Intra- and Inter-Corpus Similarity}
\label{sec:corpus_sim}

%\thai{Can we merge somehow with the previous RQ2? Or can we title "Plagiarism v.s. Intra- and Inter-Corpus Similarity"}

\begin{figure}[t]
  \centering
  \includegraphics[width=\linewidth, height=8cm, keepaspectratio]{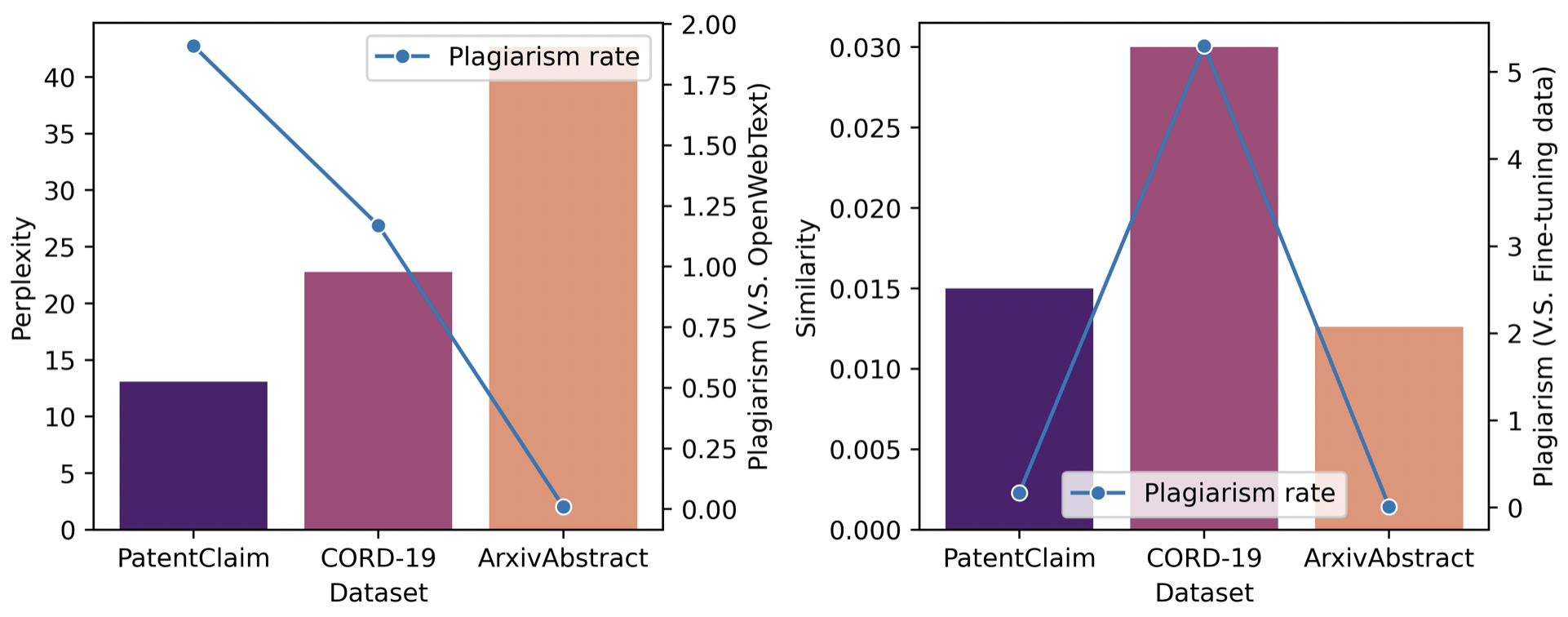}
    \caption{Perplexity (left) and similarity scores (right) of training data. %\lee{only pre-training? explain what left and right figure shows} 
    Plagiarism rate represents the average percentage of all plagiarism categories using the three decoding methods.}
    %\thai{Can we normalize the y-axis over 5000? The current scale will confuse the readers on ``similarity". And can we add another y-axis on the right of each plot to show the correlation with plagiarism?}}
    %\todo{can you change the first figure (swap Patent and CORD position but keep the color unchanged) so that the trend is clear?}
  \label{fig:perp_sim}
\end{figure}

% \usepackage{multirow}
\begin{table*}[]
{
\centering
\small
\hfill{}
\begin{tabular}{@{}ccccclccc@{}}
\toprule
\multirow{2}{*}{Model}                                                & \multirow{2}{*}{Decoding} & \multicolumn{3}{c}{\textbf{Before} Filtering Low Perplexity}       &  & \multicolumn{3}{c}{\textbf{After} Filtering Low Perplexity}             \\ \cmidrule(lr){3-5} \cmidrule(l){7-9} 
                                                                      &                           & Verbatim  & Paraphrase            & Idea                  &  & Verbatim         & Paraphrase           & Idea                 \\ \midrule
\multirow{3}{*}{\begin{tabular}[c]{@{}c@{}}Patent\\ GPT\end{tabular}} & temp*                      & 0 (0\%)   & 26 (0.52\%)         & 11 (0.22\%)           &  & 0 (0\%)          & 11 (0.22\%) & 9 (0.18\%)  \\
                                                                      & top-\emph{k}*                     & 0 (0\%)   & 109 (2.18\%) & 109 (2.18\%) &  & 0 (0\%)          & 79 (1.58\%) & 54 (1.08\%) \\
                                                                      & top-\emph{p}*                     & 0 (0\%)   & 66 (1.32\%)           & 59 (1.18\%)           &  & 0 (0\%)          & 41 (0.82\%) & 27 (0.54\%) \\ \midrule
\multirow{3}{*}{\begin{tabular}[c]{@{}c@{}}Cord19\\ GPT\end{tabular}} & temp                      & 0 (0\%)   & 7 (0.14\%)            & 6 (0.12\%)            &  & 0 (0\%)          & 4 (0.08\%)  & 1 (0.02\%) \\
                                                                      & top-\emph{k}*                     & 0 (0\%)   & 67 (1.34\%)           & 106 (1.12\%)          &  & 0 (0\%)          & 56 (1.12\%) & 36 (0.72\%) \\
                                                                      & top-\emph{p}*                     & 5 (0.1\%) & 54 (1.08\%)           & 59 (1.18\%)           &  & 0 (0\%) & 35 (0.7\%)  & 25 (0.5\%)  \\ \bottomrule
                                            %\multicolumn{8}{l}{\thai{What bold scores mean?}}
\end{tabular}
\hfill{}
}

\caption{Number (\%) of machine-generated documents w.r.t. three plagiarism types before/after removing training samples with low perplexity. The total number of generated documents for each model and decoding method is 5,000. * indicates a statistical significance ($p$ < 0.05).}
\label{tab:perp_plagiarism}
\vspace{-0.1in}
\end{table*}

\subsection{Inter-Corpus Similarity (across Datasets)}

\noindent \textbf{Method.} There are various methods to compute a corpus similarity. Generally speaking, we first transform document pairs into vectors, apply pair-wise document similarity measurements, and then aggregate them. Yet, since the size of OpenWebText is huge, it is computationally expensive to employ conventional approaches. Thus, inspired by \citet{kilgarriff1998measures} and \citet{carlini2021extracting}, we utilize perplexity measures. The \emph{perplexity} of a sequence estimates the confidence levels of an LM when predicting the inclusive tokens in a specific order. To compute the corpus similarities of pre-training and fine-tuning sets, we retrieve the perplexity of the pre-trained GPT-2 on the fine-tuning dataset. Due to the limited space, we refer the readers to the Appendix \ref{sec:appendix6} for a detailed description of perplexity calculation.

\begin{table*}[]
{
\centering
\small
\hfill{}
\begin{tabular}{@{}ccccclccc@{}}
\toprule
\multirow{2}{*}{Model}                                                & \multirow{2}{*}{Decoding} & \multicolumn{3}{c}{\textbf{Before} Filtering Similar Documents} &  & \multicolumn{3}{c}{\textbf{After} Filtering Similar Documents}                \\ \cmidrule(l){3-9} 
                                                                      &                           & Verbatim      & Paraphrase     & Idea                  &  & Verbatim             & Paraphrase            & Idea                  \\ \midrule
\multirow{3}{*}{\begin{tabular}[c]{@{}c@{}}CORD19\\ GPT\end{tabular}} & temp                      & 15 (0.3\%)    & 64 (1.28\%)    & 22 (0.44\%)  &  & 10 (0.2\%)  & 49 (0.98\%)  & 25 (0.5\%)            \\
                                                                       & top-\emph{k}*                     & 11 (0.22\%)   & 301 (6.02\%)   & 238 (4.76\%)          &  & 11 (0.22\%)          & 203 (4.06\%) & 184 (3.68\%) \\
                                                                      & top-\emph{p}*                     & 21 (0.42\%)   & 190 (3.8\%)    & 111 (2.22\%)          &  & 11 (0.22\%) & 153 (3.06\%) & 94 (1.88\%)  \\ \bottomrule
                                      %\multicolumn{8}{l}{\thai{What bold scores mean?}}
\end{tabular}
\hfill{}
}

\caption{Number (\%) of machine-generated documents w.r.t. three plagiarism types before/after removing similar training samples. The total number of generated documents for each model and decoding method is 5,000. * indicates a statistical significance ($p$ < 0.05).}
\label{tab:sim_plagiarism}
\vspace{-0.1in}
\end{table*}

%\\\vspace{0.5em}
 \vspace{0.05in}\noindent \textbf{Results.} A low perplexity implies that LM is not surprised by the sequence. In our case, the lower the perplexity score is, the more comparable a particular fine-tuned corpus is to OpenWebText. We find that a perplexity score of PatentClaim is the lowest, following Cord-19 and ArxivAbstract (Figure \ref{fig:perp_sim}). This result concurs with our initial observation where PatentGPT plagiarizes the most from OpenWebText. Subsequently, we create two versions of PatentGPT and Cord19GPT to test the effect of perplexity on plagiarism from OpenWebText. While the first is trained with a subset of fine-tuning samples excluding 30\% of the documents with the lowest perplexity, the second does not consider the perplexity. 

For a fair comparison, we maintain the same training configurations for all model pairs.\footnote{PatentGPT variations are trained on 189,000 documents for 22,000 steps, whereas Cord19 variations are trained on 140,000 documents for 40,850 steps.} Finally, we generate 5,000 documents for each model using three decoding methods and compare their plagiarism. As shown in Table \ref{tab:perp_plagiarism}, omitting low perplexity documents mitigates the intensity of plagiarism from pre-training data.\footnote{Refer to Appendix \ref{sec:statistics} for statistical testing results.}

\subsection{Intra-Corpus Similarity (within Datasets)}
\textbf{Method.} 
Here we adopt a traditional document similarity measurement to quantify inner-similarity levels of fine-tuning datasets. For each fine-tuning data, we first convert all instances into term frequency-inverse document frequency (tf-idf) vectors and then aggregate the averaged cosine similarity over all examples. 

%\\\vspace{0.5em}
 \vspace{0.05in}\noindent \textbf{Results.}
We observe that the intra-corpus similarity of Cord-19 is more than twice higher than PatentClaim and ArxivAbstract (Figure \ref{fig:perp_sim}). This result coincides with our observation in RQ2 where Cord19GPT demonstrates a heightened degree of plagiarism. Moreover, our manual inspection of verbatim plagiarism cases supports that most of them are frequently occurring substrings. For example, a part of BMJ's statement about copyright and authors’ rights\footnote{https://authors.bmj.com/policies/copyright-and-authors-rights/} appeared 588 times in the Cord-19 corpus. We further evaluate a correlation between corpus homogeneity and plagiarism by re-training two Cord19GPT models. Specifically, the former is fine-tuned with randomly selected 188,880 Cord-19 documents whereas the latter is fine-tuned using filtered Cord-19 data where 11,120 highly similar training instances (cosine similarity > 0.8) are removed. They are both trained for roughly 42,390 steps. Table \ref{tab:sim_plagiarism} supports the effectiveness of removing similar training instances in reducing plagiarism from fine-tuning data.\footnote{Refer to Appendix \ref{sec:statistics} for statistical testing results.}

\section{Findings}  
 %We elaborate major findings here.

 \vspace{0.05in}
\noindent \textbf{1. Larger LMs plagiarize more.} 
Consistent with \citet{carlini2021extracting} and \citet{carlini2022quantifying}, we find that larger GPT-2 models (large and xl) generally generate plagiarized sequences more frequently than smaller ones. Depending on the decoding approaches, however, the model size that yields the largest amount of plagiarism change: when the next token is sampled from truncated distribution, the GPT-2 large model plagiarizes the most. On the other hand, the GPT-2 xl becomes more strongly associated with plagiarism than the GPT-2 large when the temperature setting without truncation is employed. This discrepancy may be attributable to the error rates of our paraphrase and idea plagiarism detection tool. Regardless, it is evident that larger models plagiarize notably more from training data. Considering the performance improvement of LMs with larger model sizes, this finding sheds light on a trade-off between the performance and  copyright protection issues. 

\vspace{0.05in}
\noindent \textbf{2. Decoding algorithms affect plagiarism.} 
Varying effects of decoding methods and parameters on text quality and diversity have been extensively studied \cite{delucia2020decoding, basu2020mirostat}, but not from the plagiarism perspective. Particularly, top-\emph{p} sampling is reported to be the most effective decoding method in generating high-quality texts \cite{ippolito2019automatic}. Despite its efficiency in balancing quality and novelty, our analysis shows that sampling with top-\emph{p} or top-\emph{k} truncation leads to more plagiarism cases. This result shows that these popular sampling approaches still pose critical flaws because they have not been thoroughly vetted in terms of plagiarism. Thus, it is necessary to carefully choose and evaluate decoding methods not only through the lens of quality and diversity but also through the originality aspect. 

% \hfill\break
%Similarly, it is well known that decoding parameters substantially influences machine-generated texts: the novelty can be enhanced by increasing parameter values (\emph{t, k, p}), but comes at the cost of degraded quality \cite{mccoy2021much,basu2020mirostat}. \citet{radford2019language} has recommended users set \emph{t} to 1, \emph{k} to 40, and \emph{p} as values ranging from 0.8 to 1.0 because a trade-off between quality and novelty in neural texts is relatively small. Yet, these parameters have not been thoroughly vetted in terms of plagiarism. It therefore is critical to carefully reevaluate these elements not only through the lens of quality and diversity but also through plagiarism aspects. \hfill\break

% research question2
%\noindent \textbf{Fine-tuning dataset matters: Corpus similarity and homogeneity affect plagiarism.} 
\vspace{0.05in}
\noindent \textbf{3. Fine-tuning LMs matter.} 
%To the best of our knowledge, we're the first to inspect plagiarism issues of fine-tuned LMs. 
Our findings highlight that fine-tuning a model with domain-specific data can mitigate verbatim plagiarism from the pre-training dataset. Still, other types of plagiarism cases have surged, in the case of PatentGPT and Cord19GPT, alongside corpus similarity levels between pre-training and fine-tuning corpora. Moreover, we observe that models' plagiarism differs depending on similarity degrees within a fine-tuning corpus. Our research validates their relationships by comparing the rate of plagiarism before and after removing syntactically or semantically similar instances in fine-tuning data. Indeed, restricting inter- and intra-corpus similarity can reduce the frequency of all plagiarism types. This result can further be expanded as a simple solution to LMs' plagiarism issues.

%this does not influence plagiarism from the fine-tuning corpus: only the Cord19GPT demonstrates intensified degree of plagiarism  where plagiarized documents make up to 6\%. We are uncertain why Cord19GPT behaves differently, but we assume this is due to the specificity of the CORD-19 dataset. 

%Inspired by \cite{lee2021deduplicating}'s results that models tend to remember common and overlapping knowledge more easily, we hypothesize that corpus similarity between original and auxiliary corpora affects fine-tuning models' plagiarism. Indeed, our observations identify a noticeable difference in similarity levels and confirm its positive impact on plagiarism from the OpenWebText. 

\vspace{0.05in}
\noindent \textbf{4. LMs can pose privacy harms.} 
Our qualitative examination of plagiarized texts reveals that LMs expose individuals' sensitive or private data not only through verbatim plagiarism but also paraphrase and idea plagiarism. Although all identified contents were publicly available on the Web, emitting such sensitive information in the generated texts can raise a serious concern. This finding adds value to the ongoing discussion around privacy breaches from the memorization of modern LMs.

%Our research overall has raised a concern towards the growing use of a LM, considering its potential harm on both our privacy and authorship.
%Our findings add value to ongoing discussions around privacy breaches resulting from the memorization of deep neural language models. We discover multiple plagiarized examples where users' sensitive or private data such as phone number or email address is exposed. Although all identified content were publicly available on the Web, it does not give a right for LMs to reveal their personal information without consent. Our research overall has raised a concern towards the growing use of a language model, considering its potential harm on both our privacy and authorship. 

\section{Discussion and Ethics} 

\noindent \textbf{Discussion.}
In this work, we develop a novel pipeline for investigating LMs' plagiarism in text generation processes and characterize a shift in plagiarism rates resulting from three attributes (i.e., model size, decoding methods, and corpus similarities). The datasets utilized to train the models are the subject of this study.
We use GPT-2 as a representative LM to study because it is one of the most downloaded LMs from Hugging Face at the end of 2022,\footnote{\url{https://huggingface.co/models?sort=downloads}} and its reproduced training corpus is publicly accessible (which is a necessary condition to study the plagiarism of LMs). However, different LMs may demonstrate different patterns of plagiarism, and thus our results may not directly generalize to other LMs, including more recent LMs such as GPT-3 or BLOOM. Future work can revisit the proposed research questions against more diverse or modern LMs. 

%Nonetheless, these models have not released their training corpora, making it difficult to run our experiments. 

% more recent models such as GPT3 have not disclosed their training corpus so we couldn’t experiment, (2) although GPT2 was proposed a few years ago, it’s still one of the most popular general-purpose LMs, (3) we strongly believe our findings can similarly occur in more recent models too, which future work would verify. 

In addition, automatic plagiarism detectors are known to have many failure modes (both in false negatives and false positives) \cite{weber2019plagiarism}. Our plagiarism detection pipeline of Section \ref{sec:plagiarism_framework} is no exception. However, achieving a high precision with a low recall is not a major issue in our problem domain, as we focus on demonstrating the lower-bound of the plagiarism vulnerability in LMs (and in reality, there are likely to be many more plagiarism cases that we missed to detect due to low recalls). Likewise, prior memorization works \cite{carlini2021extracting, kandpal2022deduplicating} documented the lower-bound of the plagiarism susceptibility and showed a small number of memorized instances. Regardless, they were effective in inspiring others to continue exploring this important phenomenon. As a result, we hope that our current finding becomes useful to stimulate and raise public awareness about the plagiarism behavior of popular LMs like GPT-2.

We also stress that distinguishing whether a reproduction of training datasets is a positive attribute of LM or not is beyond the scope of this work. It is highly context-dependent \cite{lee2021deduplicating}, and thus necessitates more sophisticated methods to disentangle. In our experiments, we treat all instances of LM-generated texts that reiterate training examples as ``problematic", as the fine-tuning datasets we analyzed are in academic and legal contexts where originality is valued.
%
%Whether a reproduction of training datasets is a positive attribute of LM, or dangerous and undesirable, is context-dependent \cite{lee2021deduplicating}. For instance, in general, the memorized content can be unwanted (when it contains someone's private information), neutral (when it includes common phrases) and desired (when it's factual information). Yet, due to subjectivity, distinguishing them is extremely complex. Moreover, this boundary may evolve depending on the model's downstream applications: if the model is employed for creative writings, a majority of memorized cases are likely to violate societal or ethical norms because they cannot cite the source as humans do. Because the datasets we analyzed are in academic and legal contexts where originality is valued, we treat all instances of the LM generated text that reiterate training examples as problematic in our experiments.
%

Ultimately, a primary purpose of the exploration of the intra- and inter-corpus similarity in models' authorship violation is to support our hypotheses and further motivate researchers to take this into account when developing new LMs or fine-tuning current ones. Yet, the current approach fails to completely eradicate plagiarism occurrences. 

\vspace{0.05in}
\noindent \textbf{Ethics.}
 Data and code, involving plagiarized texts we identified throughout this research, are available to the research community. Due to the inclusion of individuals' personal data in generated texts, we employed data anonymization techniques prior to distribution. %Given that OpenWebText is crawled from Reddit, we will transform the self-selected user names into arbitrary numeric values and filter personally identifiable information using Microsoft's Presidio Anonymizer.\footnote{\url{https://microsoft.github.io/presidio/anonymizer/}} 
Specifically, we filtered PII such as name, email address, and phone number using Microsoft's Presidio Anonymizer.\footnote{\url{https://microsoft.github.io/presidio/anonymizer/}} 
We recommend that artificial documents generated by fine-tuned GPT-2 be utilized strictly for research purposes.

\section{Conclusion}
Our work presents the first holistic and empirical analyses of plagiarism in LMs by constructing a pipeline for the automatic identification of plagiarized content. We conclude that GPT-2 can exploit and reuse words, sentences, and even core ideas (that are originally included in OpenWebText, a pre-training corpus) in the generated texts. Further, this tendency is prone to exacerbate as the model size increases or certain decoding algorithms are employed. We also discover that untangling corpus similarity and homogeneity can help alleviate plagiarism rates by GPT-2.
This is the first study to analyze text generation outputs through the lens of plagiarism. Although the goal of a supervised machine learning system is to learn to mimic the distribution of its training data, we deem it crucial for model users and designers to recognize the observed phenomena. The vulnerability of models to plagiarism can adversely impact societal and ethical norms, particularly in literary disciplines that are intimately connected to creativity and originality. Therefore, we recommend researchers carefully assess the model's intended usage and evaluate its robustness before deployment.

\begin{acks}
This work was in part supported by NSF awards \#1934782 and \#2114824.
\end{acks}


\bibliographystyle{ACM-Reference-Format}
\bibliography{sample-base}

\newcolumntype{P}[1]{>{\centering\arraybackslash}p{#1}}

%\newcolumntype{C}{>{\Centering\arraybackslash}X} % centered "X" column
\begin{table*}[!htbp]
 \centering
{\footnotesize
\hfill{}
\begin{tabular}{P{0.09\textwidth}|P{0.41\textwidth}P{0.41\textwidth}}
%\begin{xtabular}{|p{0.48\textwidth}|p{0.48\textwidth}|}
\toprule
Type & Machine-Written Text & Training Text \\
\cmidrule(lr){1-3}
\multirow{3}*{Verbatim} &  \MyDBox[yellow!30]{Unexpected Error An unexpected error occurred. {\color{olive}[...]} "facebookAppID":***,"allow\_select":true,"allow\_filter":true,"allow\_sheetlink":true {\color{olive}[...]}} (\textit{\textbf{Author: GPT-2}})
 &  \MyDBox[yellow!30]{Unexpected Error An unexpected error occurred. {\color{olive}[...]} "facebookAppID":***,"allow\_select":true,"allow\_filter":true,"allow\_sheetlink":true {\color{olive}[...]}}  \\  
\cmidrule(lr){1-3}
\multirow{3}*{Verbatim} & \MyDBox[yellow!30]{it reminded me of a feeling I've had right there on that road before. It reminded me of all the times that people have come out to support the blockade and stood together to make sure those trees stay standing. {\color{olive}[...]}} (\textit{\textbf{Author: GPT-2}})
&  \MyDBox[yellow!30]{it reminded me of a feeling I've had right there on that road before. It reminded me of all the times that people have come out to support the blockade and stood together to make sure those trees stay standing. {\color{olive}[...]}} \\ 
\cmidrule(lr){1-3}
\multirow{3}*{Verbatim} & \MyDBox[yellow!30]{I, the Submitting Author has the right to grant and does grant on behalf of all authors of the Work (as defined in the below author licence), an exclusive licence and/or a non-exclusive licence for contributions from authors who are: i) UK Crown employees; ii) where BMJ has agreed a CC-BY licence shall apply, and/or iii) in accordance with the terms applicable for US Federal Government officers or employees acting as part of their official duties; {\color{olive}[...]}}(\textit{\textbf{Author: Cord19GPT}}) 
&  \MyDBox[yellow!30]{I, the Submitting Author has the right to grant and does grant on behalf of all authors of the Work (as defined in the below author licence), an exclusive licence and/or a non-exclusive licence for contributions from authors who are: i) UK Crown employees; ii) where BMJ has agreed a CC-BY licence shall apply, and/or iii) in accordance with the terms applicable for US Federal Government officers or employees acting as part of their official duties; {\color{olive}[...]}} \\ 
\cmidrule(lr){1-3}
\multirow{6}*{Paraphrase} & \MyDBox[yellow!30]{REUTERS/Kevin Lamarque U.S. President Donald Trump} and \MyDBox[yellow!30]{First Lady Melania Trump},  with \MyDBox[yellow!30]{their son Barron}, arrive for \MyDBox[orange!30]{a New Year's Eve} party at his \MyDBox[yellow!30]{Mar-a-Lago club in Palm Beach, Florida}, U.S. December 31, 2017. {\color{olive}[...]} (\textit{\textbf{Author: GPT-2}}) & \MyDBox[yellow!30]{REUTERS/Kevin Lamarque U.S. President Donald Trump}, \MyDBox[yellow!30]{First Lady Melania Trump} and \MyDBox[yellow!30]{their son Barron} while aboard Air Force One on their way to Florida, \MyDBox[yellow!30]{Mar-a-Lago in Palm Beach, Florida} to spend \MyDBox[orange!30]{the holiday} at Trump International Golf Club Mar-a-Lago. {\color{olive}[...]} \\
\cmidrule(lr){1-3}
\multirow{5}*{Paraphrase} & The development of \MyDBox[yellow!30]{natural killer cells (NK cells)} is \MyDBox[yellow!30]{an important} element in \MyDBox[orange!30]{the immune system as it provides the first line of defense against diverse pathogens.} (\textit{\textbf{Author: Cord19GPT}}) & \MyDBox[yellow!30]{Natural killer (NK) cells} are a type of innate lymphoid cell that plays \MyDBox[yellow!30]{an important} role \MyDBox[orange!30]{in the first line of immune defense against any viral infection}, including COVID-19. \\
\cmidrule(lr){1-3}
\multirow{5}*{Paraphrase} & A system, comprising: a sense circuit for receiving an electrical {\color{olive}[...]} \MyDBox[orange!30]{and a digital compensator coupled with the sense circuit and for receiving the output value from the decision circuit and generating a compensation value in accordance with the output value} {\color{olive}[...]}  (\textit{\textbf{Author: PatentGPT}}) &  Apple's First Claim: A touch surface device, comprising: a touch-sensitive panel {\color{olive}[...]}
\MyDBox[orange!30]{and a sensing circuit coupled to the compensation circuit, the sensing circuit configured for receiving the compensated output signal.}
\\
\cmidrule(lr){1-3}
\multirow{9}*{Idea} &  A method for testing electrical connections, comprising: {\color{olive}[...]} providing an electric voltage and an \MyDBox[yellow!30]{electric current} to an electrical contact on the test element to \MyDBox[orange!30]{transfer the electrical conductivity of the line to ground; wherein the measuring is carried out with the electric current flowing from the electrical contact on the test element through the electric current to the ground;} {\color{olive}[...]} (\textit{\textbf{Author: PatentGPT}}) & The energy passing between elements A and B is in the form of an \MyDBox[yellow!30]{electric current} \MyDBox[orange!30]{through the earth between the two ground connections.} \\
\cmidrule(lr){1-3}
\multirow{9}*{Idea} &  A control system comprising: \MyDBox[orange!30]{a processor configured to execute an operation on} \MyDBox[yellow!30]{a memory} and \MyDBox[orange!30]{to output an instruction stream having a plurality of executable instructions}, wherein the output of the plurality of executable instructions is selectively selectable {\color{olive}[...]}; and \MyDBox[orange!30]{a storage device storing a plurality of items of a control structure, each of the control structures containing executable instructions, which when executed by the processor,} \MyDBox[orange!30]{cause the processor to perform} {\color{olive}[...]} (\textit{\textbf{Author: PatentGPT}})
&  The system also may comprise \MyDBox[yellow!30]{a memory} \MyDBox[orange!30]{having stored thereon instructions that, upon execution by the at least one processor}, \MyDBox[orange!30]{cause the system to perform} {\color{olive}[...]}  \\
\cmidrule(lr){1-3}
\multirow{9}*{Idea} & \MyDBox[yellow!30]{Symptoms of COVID-19} infections are relatively mild, such as fever, dry cough, \MyDBox[yellow!30]{headache, diarrhea}, dyspnoea, body ache, \MyDBox[yellow!30]{myalgia} and sometimes headache. In some infected patients, however, the infection is more rapid and severe with \MyDBox[yellow!30]{fever}, dyspnoea, \MyDBox[yellow!30]{shortness of breath, cough} and other non-specific symptoms such as \MyDBox[yellow!30]{sore throat}, \MyDBox[orange!30]{runny nose}, dry throat and sputum production. {\color{olive}[...]} Several factors are strongly associated with mortality in the SARS-CoV-2 outbreak. {\color{olive}[...]} and \MyDBox[orange!30]{comorbidities such as} \MyDBox[yellow!30]{hypertension, obesity,} \MyDBox[orange!30]{chronic lung disease}, \MyDBox[yellow!30]{obesity and diabetes}. (\textit{\textbf{Author: Cord19GPT}})
&
The most common \MyDBox[yellow!30]{symptoms of COVID-19} are \MyDBox[yellow!30]{headache}, loss of smell, \MyDBox[orange!30]{nasal congestion}, \MyDBox[yellow!30]{cough}, asthenia, \MyDBox[yellow!30]{myalgia}, rhinorrhea, \MyDBox[yellow!30]{sore throat, fever, shortness of breath}, nausea or vomiting, and \MyDBox[yellow!30]{diarrhea} [2, 3] . Commonly reported \MyDBox[orange!30]{comorbidities of COVID-19 are} \MyDBox[yellow!30]{hypertension, obesity, diabetes,} and \MyDBox[orange!30]{cardiovascular disease [4].}  \\
\bottomrule
\end{tabular}
}
\caption{Examples of plagiarism identified in texts written by GPT-2 and its training set. Duplicated texts are highlighted in \MyDBox[yellow!30]{yellow}, and words/phrases that contain similar meaning with minimal text overlaps are highlighted in \MyDBox[orange!30]{orange}. {\color{olive}[...]} indicates the texts omitted for brevity. Personally identifiable information (PII) was masked as ***.}
\label{tab:real_examples2}
\end{table*}

\appendix

% table
\begin{table*}[!htbp]{
\centering
\small
\hfill{}
\begin{tabular}{ccc}
\hline
Model Name          & Training Steps & Training / Test Loss \\ \hline
ArXivAbstractGPT  & 30,000            & 2.48 / 2.83          \\ 
Cord19GPT       & 44,000            & 2.6 / 2.68           \\ 
PatentGPT     & 32,300        & 1.65 / 1.87        \\ \hline
\end{tabular}
\hfill{}
}
\caption{Fine-tuning configurations}
\label{tab:finetuning_conf}
\end{table*}

\begin{table*}[!htbp]
\centering
\small
\hfill{}
\begin{tabular}{@{}cccc@{}}
\toprule
Model                                                                 & Decoding & Plagiarized Document \# (before filtering vs. after filtering ) & $p$                 \\ \midrule
\multirow{3}{*}{\begin{tabular}[c]{@{}c@{}}Patent\\ GPT\end{tabular}} & temp     & 37 vs. 20                                                       & 0.002           \\
                                                                      & top-\emph{k}    & 218 vs. 133                                                     & \textless 0.00001 \\
                                                                      & top-\emph{p}    & 125 vs. 86                                                      & 0.007            \\ \midrule
\multirow{3}{*}{\begin{tabular}[c]{@{}c@{}}Cord19\\ GPT\end{tabular}} & temp     & 13 vs. 5                                                        & 0.059            \\
                                                                      & top-\emph{k}    & 173 vs. 92                                                      & \textless 0.00001 \\
                                                                      & top-\emph{p}    & 118 vs. 60                                                      & 0.00002           \\ \midrule
                                                                      \midrule
\multirow{3}{*}{\begin{tabular}[c]{@{}c@{}}Cord19\\ GPT\end{tabular}} & temp     & 101 vs. 84                                                      & 0.207             \\
                                                                      & top-\emph{k}    & 550 vs. 398                                                     & \textless 0.00001 \\
                                                                      & top-\emph{p}    & 322 vs. 258                                                    & 0.006            \\ \bottomrule
\end{tabular}
\hfill{}
\caption{Statistical results of the chi-squared test. The first result regarding Cord19GPT is for perplexity, whereas the second one is for document similarity.}
\label{tab:p-values}
\end{table*}


\section{Evaluation Data for Our Plagiarism Detection Pipeline}
\label{sec:appendix7}

We use two corpora with plagiarism labels to measure the precision and recall scores of our proposed pipeline described in Section \ref{sec:plagiarism_framework}. The first dataset (denoted as \emph{PanDataset}) is originally introduced as a test set for the fifth international competition on plagiarism detection at PAN 2013.\footnote{\url{https://pan.webis.de/clef13/pan13-web/text-alignment.html}} It contains in total 3,170 source documents and 1,827 suspicious documents where 1,001 document pairs are without plagiarism and 1,001 pairs are affiliated with verbatim plagiarism. In order to automatically create document pairs for paraphrase plagiarism, the organizers applied machine-driven approaches such as randomly replacing words based on a synonym database like WordNet or back-translating sentences with existing translation models (e.g., Google Translate\footnote{\url{http://translate.google.com}}) using source documents. This resulted in 2,002 pairs. Similarly, 1,186 summary plagiarism cases are generated by existing text summarization models.

Given that PanDataset may exhibit different characteristics from GPT-2 generated texts, we consider a subset of OpenWebText as source documents, create suspicious documents, and use the pairs as the second dataset (denoted as \emph{GptPlagiarismDataset}). More specifically, we construct 1,000 document pairs for verbatim plagiarism by extracting 500 character-long texts within source documents and using them as suspicious documents. For paraphrase plagiarism, we randomly select 5 sentences from 1,000 source documents and employ Facebook FAIR’s WMT19 transformer \cite{ng2019facebook} for back translation (English->German->English). Lastly, 1,000 document pairs for summary plagiarism are created by two summarization models. We first shorten the lengths of source documents with a BERT-based extractive summarization model \cite{miller2019leveraging} and then transformed them into meaningful summaries using T5 transformer \cite{raffel2020exploring} for abstractive summarization. This enables us to create more sophisticated summaries with minimal overlapping strings.

\section{Details on Fine-tuning Configurations}
\label{sec:appendix5}

Our experimental environment is based on a Google Colab Pro+ with Tesla V100-SXM2-16GB and 55 GB of RAM. For fine-tuning, we utilize a Python package called GPT-2-simple.\footnote{\url{https://github.com/minimaxir/gpt-2-simple}} We maintain hyperparameters that are suggested in public repositories: learning rate as 1e-4, temperature as 1.0, top-\emph{k} as 40, and batch size as 1. The ratio of training and validation sets is 8:2. To prevent the model from overfitting, we stop training processes when a gap between training and test losses reaches over 20\% of training loss. Table \ref{tab:finetuning_conf} illustrates their fine-tuning configurations. Fine-tuning one model for 10,000 steps approximately takes 5 hours.

\section{LM Perplexity Calculation}
\label{sec:appendix6}
Perplexity is defined as the exponentiation of the cross-entropy between the data and LM predictions. Given a tokenized sequence \(X = (x_{0}, x_{1}, x_{2}...x_{n})\), the perplexity of \(X\) can be calculated by:
\[perp(X) = exp\left \{ -\frac{1}{m} \sum_{n}^{m} logf_{\theta } (x_{n}| x_{\leq n-1} ) \right \} \]

\noindent where \(logf_{\theta } (x_{n}| x_{\leq n-1} )\) is the log-likelihood of the \(n\)th token conditioned on the preceding tokens. Following the guideline provided by Huggingface,\footnote{\url{https://huggingface.co/docs/transformers/perplexity}} we rely on a strided sliding-window technique, which entails moving the context window repeatedly so that the model has a broader context when making each prediction. Here a window size is a hyper-parameter we can adjust. To retrieve one aggregated perplexity that represents the whole instances, we first append all documents with newlines and then set the window size as 512. For an individual document perplexity calculation of the Cord-19 dataset, we reduce the window size to 50 since we do not append all documents this time, and many Cord-19 documents tend to be shorter than 512 tokens.

\section{Statistical Testing of Filtering}
\label{sec:statistics}

We perform the Pearson’s chi-squared test \cite{plackett1983karl} to verify the statistical significance of the observed gap between before and after filtering low-perplexity and similar documents. The test is used to determine whether there is a statistically significant difference between the expected frequencies and the observed frequencies. Here we treat plagiarism as a binary variable (no plagiarism vs. plagiarism) and count the total number of documents accordingly. For plagiarized document count, we do not distinguish plagiarism types. Table \ref{tab:p-values} shows the results of the chi-squared test. Most of our experiments except for Cord19GPT's temperature setting are found to be statistically meaningful.

\section{Plagiarized Text Examples}
\label{sec:appendix4}
We present several examples of verbatim, paraphrase, and idea plagiarism from both pre-trained and fine-tuned models (Table \ref{tab:real_examples2}). For verbatim plagiarism, we identify cases where social media's app ID and its metadata are memorized, as well as an individual's writing. We also frequently find a paragraph related to journals' copyright and authors’ rights as verbatim plagiarism from the model trained with academic papers. Examples associated with paraphrase plagiarism, especially those authored by GPT-2 and Cord19GPT, demonstrate models' abilities in delivering factual information in a different syntactic form without proper references. PatentGPT's plagiarism cases tend to mimic patent data by rephrasing and elaborating on the described processes created by original patent owners.

\end{document}